\documentclass[10pt,twocolumn,letterpaper]{article}

\pdfpageattr{/Group <</S /Transparency /I true /CS /DeviceRGB>>} 

\usepackage{authblk}
\usepackage{varwidth}
\usepackage{lmodern}
\usepackage{minibox}
\usepackage{comment}

\makeatletter
\renewcommand\AB@affilsepx{\hspace{1.2cm}}
\makeatother

\usepackage[dvipsnames,svgnames,x11names,table, dvipsnames]{xcolor}
\usepackage[markup=underlined]{changes}
\definecolor{blueNote}{HTML}{2B619D}
\definechangesauthor[color=blueNote,name=\textbf{Alejandro}]{AC}
\definechangesauthor[color=red,name=\textbf{Mariella}]{MD}
\usepackage{todonotes}
\makeatletter
\setremarkmarkup{\todo[color=Changes@Color#1!20,size=\scriptsize]{#1: #2}}
\makeatother

\usepackage{iccv}
\usepackage[paperheight=279.4mm,paperwidth=215.9mm]{geometry}
\newdimen\extramargin
\usepackage{times}
\usepackage{amsmath}
\usepackage{amssymb}
\usepackage{booktabs}
\usepackage{rotating}
\usepackage{multirow}

\usepackage{epsfig}
\usepackage{graphicx}
\graphicspath{{images/}}
\DeclareGraphicsExtensions{.eps}
\DeclareGraphicsExtensions{.png}
\DeclareGraphicsExtensions{.jpg}

\usepackage[pagebackref=true,breaklinks=true,letterpaper=true,colorlinks,bookmarks=false]{hyperref}
\usepackage{subcaption}
\iccvfinalcopy 

\def\iccvPaperID{} 

\ificcvfinal\fi
\begin{document}

\title{Social Style Characterization from Egocentric Photo-streams}
\author{Maedeh Aghaei\textsuperscript{1,2}, Mariella Dimiccoli\textsuperscript{1,2},  Cristian Canton Ferrer\textsuperscript{3}, Petia Radeva\textsuperscript{1,2}\\
\textsuperscript{1}University of Barcelona, Mathematics and Computer Science Department, Barcelona, Spain\\
\textsuperscript{2}Computer Vision Center, Universitat Aut\'onoma de Barcelona, Cerdanyola del Valles, Spain\\
\textsuperscript{3}Microsoft Research, Redmond (WA), USA
}


\maketitle

\begin{abstract}
\end{abstract}
This paper proposes a system for automatic social pattern characterization using a wearable photo-camera. The proposed pipeline consists of three major steps. First, detection of people with whom the camera wearer interacts and, second, categorization of the detected social interactions into formal and informal.  These two steps act at event-level where each potential social event is modeled as a multi-dimensional time-series, whose dimensions correspond to a set of relevant features for each task, and a LSTM network is employed for time-series classification. In the last step, recurrences of the same person across the whole set of social interactions are clustered to achieve a comprehensive understanding of the diversity and frequency of the social relations of the user. Experiments over a dataset acquired by a user wearing a photo-camera during a month show promising results on the task of social pattern characterization from egocentric photo-streams.

\section{Introduction}

\par Automatic analysis of data collected by wearable cameras has drawn the attention of researchers in computer vision \cite{bolanos2017toward} where social interaction analysis in particular has been an active topic of study \cite{aghaei2015towards,aghaei2016whom,alletto2015understanding,fathi2012social,narayan2014action,yang2016wearable}. 

\begin{figure*}[!t]
\centering
\includegraphics[height=6cm]{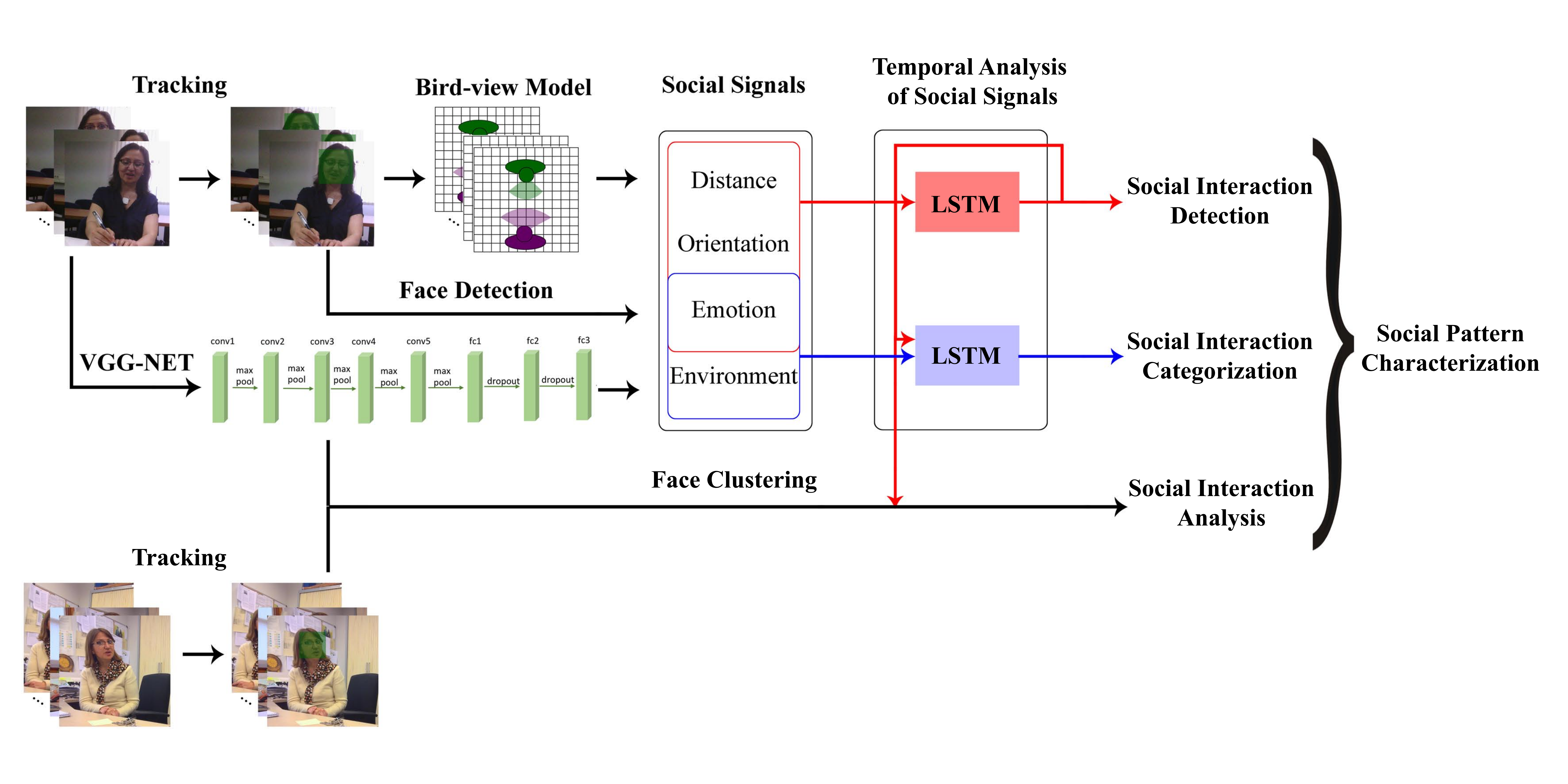}
\caption{The proposed pipeline.}
\label{fig:pipeline}
\end{figure*}

\par In this paper, we build upon our previous work \cite{aghaei2016whom} going beyond social interaction detection in egocentric photo-streams. The proposed pipeline, see Fig.\ref{fig:pipeline}, suggests firstly, to study a wider set of features for social interaction detection and secondly, to categorize the detected social interactions into two broad categories of \textit{meetings} as a special type of social interactions: \textit{formal} and \textit{informal} \cite{xiong2005meeting}. Our hypothesis is that to detect and categorize social interactions, analysis of combination of environmental features and social signals transmitted by the visible people in the scene, as well as their evolution over time is required. Eventually, social pattern characterization of the user comes naturally as the result of discovery of recurring people in the dataset and quantifying the frequency, the diversity and the type of the occurred social interactions with different individuals. Ideally, employing the entire proposed pipeline in this work, we would like to be able to answer questions such as: \textit{How often does the user engage in social interactions?} \textit{With whom does the user interact most often?} \textit{Are the interactions with this person mostly formal or informal?}\textit{How often does the user see a specific person?} 
\subsection{Social Interaction Detection}

\par Following the methodology described in \cite{aghaei2016whom}, we first segment an egocentric photo-stream into individual events \cite{dimiccoli2016sr} and select potential social events among them which are the segments with high density of appearing people. In each social event, faces are tracked by applying a multi-face tracking algorithm \cite{aghaei2016multi}. Later, the problem of social interaction detection for each tracked person is formulated as a binary time-series classification (interacting vs. no-interacting) where the time-series dimension corresponds to the number of selected social signals to describe a social interaction. In addition to the distance $(\varphi_d)$ and face orientation in terms of yaw $(\varphi_z)$ of individuals with regard to the camera-wearer proposed in \cite{aghaei2016whom}, in this work we explore the impact of face orientation also in terms of pitch $(\varphi_y)$ and roll $(\varphi_x)$ as well as of facial expressions $(\varphi_e)$. Facial expressions are represented as a vector of probabilities for each of 8 different facial expressions associated to emotions in the occidental culture \cite{BarsoumICME2016}. For a given person $p_i$, the index of the dominant facial expression as $ \displaystyle \varphi_e= \arg\max_{k \in 1,...,8} e_k(p_i)$, is considered as the facial expression value. The complete set of features is a $5-$dimensional time-series representing the time-evolution of the $j$-th interaction features over time, separately extracted for each tracked face:
\begin{equation}
\varphi_{detection}^{\tau}=(\varphi^{\tau}_d, \varphi^{\tau}_z, \varphi^{\tau}_y, \varphi^{\tau}_x, \varphi^{\tau}_e), \; \tau=1,2,\ldots
\end{equation}

\subsection{Social Interaction Categorization}
\par The sociological definition of formal and informal meetings, as the two broad categories of social interactions \cite{xiong2005meeting} from the computer vision perspective, suggests that environmental features and facial expression of people show discriminative power in meeting categorization. 

\textbf{Environmental features:} Each component of the feature vector extracted from CNN carries some semantic content, which can be considered as a good representative of the environment in an image. To reduce the curse of dimensionality of the CNN feature vector (4096D), an approach to re-writes CNN features to discrete words is applied \cite{amato2016large}. Later, PCA is applied to keep the 95\% of the most important information of the resulting sparse matrix which leads to a 35-dimensional feature vector as $\varphi_{g} \in R^{35}$. 

\textbf{Facial expression:} Facial expression features in this task are extracted as the mean of facial expressions of the total number of $n$ people detected in one frame of a sequence: $\displaystyle \varphi_{e,ind}= \frac{1}{n} \sum_{i = 1}^n e_{ind}(p_i), ind = 1, \ldots, 8$.

Our  approach  takes into account the temporal evolution of both environmental and facial expression features by modeling them as multi-dimensional time-series as $ \varphi_{categorization}^{\tau} \in R^{43} =(\varphi^{\tau}_{g}, \varphi^{\tau}_e), \; \tau=1,2,\ldots $, and relies on the LSTM for binary classification of each time-series into a formal or  an informal meeting.

\section{Social Pattern Characterization}

\subsection{Generic Interaction Characterization}
\label{sec:beyond_generic}

Characterizing the social pattern of an individual demands social interaction analysis of the user across several events during a long period of time and implies the ability of defining the nature of social interactions of the user from various temporal and social aspects. For this purpose,  we define three concepts for characterizing social interactions namely, \textit{frequency},  \textit{diversity}, and \textit{duration}.

\textbf{Frequency:} Is defined as the rate of formal (informal) interactions $I$ of a person normalized by the total number of interactions:
$F_{f (inf)}=\#{I_{f (inf) \;} }/\#{days}$

\textbf{Diversity:} Demonstrates how diverse are social interactions of a person. The term is defined as the exponential of the Shannon entropy calculated with natural logarithms, namely: 
$D =1/2 \exp \left(-\sum _{i\in\{f,inf\}}A_{i}\ln(A_{i})\right)$
where $A$ indicates whether the majority of social interactions of a person are formal (informal), respectively:
$A_{f (inf)}=\#{I_{f (inf)} \; }/\#{I\; }$. Note that when the person has the same number of formal and informal interactions (i.e. $A_{formal}=A_{informal}=0.5$),  $D=1$.

\textbf{Duration:} Is the longitude of a social interaction, defined as $L(i)$ for each social interaction $i$ of the user, it is proportional to the longitude of the sequence corresponding to that social interaction, say  $L(i) = \mathcal{T}(i)r$, where $\mathcal{T}(i)$ is the number of frames of $i$-th interaction and $r$ is the frame rate of the camera.

\subsection{Person-wise Interaction Characterization}
Person-specific social interaction characterization implies characterization of the social interactions of the user with one specific person.  For this purpose, firstly all the interactions of the user with a certain person are required to be localized. To this goal, a face clustering method adapted for egocentric photo-streams \cite{aghaei2017all} is employed, which essentially achieves the desired goal through discovery of various appearances of the same person among all the social events of the user.

Let  $\mathcal{C} = \left \{c_j\right \}$, ${j = 1, \ldots, J}$ be the set of clusters obtained by applying the face-set clustering method  on the detected interacting prototypes, where $J$ ideally corresponds to the total number of people who appeared in all social events of the user. Each cluster, $c_j$, ideally contains all the different appearances of the person $p_j$ across different social events, and $|c_j|$ is the cardinality of $c_j$ which demonstrates the number of social interactions events of the user with the person $p_j$ during the observation period. As the employed clustering method as well as the proposed method for social interaction detection and categorization act at sequence-level, inferring the interaction state of each sequence inside a cluster is straightforward. Person-wise interaction characterization of the user can be computed similar to the generic manner \ref{sec:beyond_generic}, but restricted to the interactions considered to the ones with the person of interest.

\section{Experimental Results}
\begin{table*}[!t]
\centering
\caption{EgoSocialStyle dataset}
\label{table:dataset}
\begin{tabular}{cccccccccc}
\hline
                                  
\begin{turn}{40}\#\end{turn}                                    &\begin{turn}{40}Users\end{turn}        & \begin{turn}{40}Days \end{turn}           & \begin{turn}{40}Images\end{turn}       & \begin{turn}{40}\begin{tabular}[c]{@{}l@{}}Social\\ Images\end{tabular} \end{turn}           & \begin{turn}{40}People \end{turn}       & \begin{turn}{40}Sequences\end{turn}       & \begin{turn}{40}Prototypes\end{turn}      & \begin{turn}{40}Interacting\end{turn}     & \begin{turn}{40}Formal\end{turn}         \\ \hline
{Train} & {8} & {100} & {100,000} & {3,000} & {62} & {106} & {132} & {102} & {42} \\ 
{Test}  & {1} & {30}  & {25,200}  & {2,639}  & {40} & {113} & {172} & {130} & {25} \\ \hline
\end{tabular}
\end{table*}

\begin{table*}[!t]
\centering
\caption{Social pattern characterization results}
\begin{tabular}{@{}ccccccc@{}}
\toprule
                & F-Formal & F-Informal & A-Formal & A-Informal & D & L     \\ \midrule
Generic         & 0.83             & 2.50               & 0.25                & 0.75                  & 0.87      & 25.19±1.32   \\
Person-specific & 0.25             & 1.00               & 0.20                & 0.80                  & 0.59      & 18.80 ± 0.96 \\ \bottomrule
\end{tabular}
\label{table:characterization}
\end{table*}

The proposed pipeline is evaluated over a publicly available egocentric photo-stream dataset where 8 people participated in acquiring the training set and test set is acquired by one person during one month period (Table \ref{table:dataset}).

For social interaction detection, four set of settings are explored:

\textbf{SID1:} Distance + Yaw

\textbf{SID2:} Distance + Yaw + Pitch + Roll

\textbf{SID3:} Distance + Yaw + Facial expression  

\textbf{SID4:} Distance + Yaw + Pitch + Roll + Facial expressions

\begin{table}[!t]
\caption{Social interaction detection results}
\label{tab:detection} 
\begin{tabular}{@{}cccccc@{}}
\toprule
                                       & ego-HVFF          & SID1         & SID2         & SID3         & SID4         \\ \midrule
{P}  & {82.75\%} & {80.76\%} & {88.49\%} & {88.59\%} & {\textbf{91.66\%}} \\  
{R}  & {55.81\%} & {64.61\%} & {76.92\%} & {77.69\%} & {\textbf{84.61\%}} \\  
{A}  & {58.38\%} & {61.62\%} & {75.00\%} & {75.58\%} & {\textbf{82.55\%}} \\ \bottomrule
\end{tabular}
\end{table}

SID1 is the baseline setting in which only presented features in our previous work \cite{aghaei2016whom} are studied.
 In SID2, pitch and roll in addition to yaw as the main indicator of face orientation in previous works are studied.
SID3 follows the same pattern as SID1, but includes facial expression features as well to observe the effect of facial expressions in addition to commonly studied features for social interaction detection. Finally, SID4 includes all the discussed features for social interaction detection analysis.

In Table \ref{tab:detection}, we report the obtained precision, recall and accuracy values for each settings. Besides, we also compared our obtained results with the ego-HVFF model \cite{aghaei2015towards} as the unique method amongst state-of-the-art methods suitable for social interaction detection in egocentric photo-streams. The best obtained results, in all terms belong to the SID4 setting  containing all the proposed features (distance, yaw, pitch, roll, facial expressions). 

\begin{table}[!t]
\caption{Social interaction categorization results}
\begin{tabular}{@{}cccccc@{}}
\toprule
                                   & HM-SVM    & VGG-FT         & SIC1         & SIC2         & SIC3                   \\ \midrule
{P} & {76.82\%} & {86.81\%}  & {87.91\%} & {89.01\%} & {\textbf{91.48\%}}  \\  
{R} & {63.65\%} & {89.77\%} & {90.90\%} & {92.04\%} & {\textbf{97.72\%}}  \\  
{A} & {64.87\%} & {82.30\%} & {83.18\%} & {84.95\%} & {\textbf{91.15\%}} \\ \bottomrule
\end{tabular} 
\label{tab:categorization}
\end{table}

For social interaction categorization, the following settings are considered for the temporal analysis:

\textbf{SIC1:} Environmental (VGG)

\textbf{SIC2:} Environmental (VGG-finetuned)

\textbf{SIC3:} Environmental (VGG-finetuned) + Facial expressions


We assume that global features of an event, namely environmental features, have the greatest impact in the categorization of it. Therefore, the first setting (SIC1) studies only environmental features which are extracted from the last fully connected layer of VGGNet trained over the Imagenet and preprocessed. In SIC2, the environmental features are extracted in the same manner as SIC1, but from the fine-tuned VGGNet over the training set of the proposed dataset in this work. Fine-tuning the network is achieved through instantiation of the convolutional part of the model up to the fully-connected layers and then training fully-connected layers on the photos of the training set which ideally leads to better representation of the desired classification task. SIC3 explores jointly the effect of facial expressions as well as the environmental features. 

Obtained results of this task are reported in Table \ref{tab:categorization}. We also reported results of comparing our results with state-of-the-art model HM-SVM \cite{yang2016wearable}, which employs HMM to model \textit{interaction features}, being SIC3 setting, and SVM to classify them. We also compared LSTM with CNN for frame-level classification. Quantitative results suggest the sequence-level analysis using LSTM performs better in modeling this task than frame-level analysis and LSTM provides better modeling of the problem than HMM. Moreover, on the proposed dataset, total number of 83 clusters are obtained, which is almost the double as size of the total number of prototypes in the test set. The largest cluster contains 77 number of faces of a same person, from 5 number of sequences in various social events, where 4 times out of these encounters occurred during informal meetings. 

\section{Conclusions}
\label{sec:conclusion}
In this work, we proposed a complete pipeline for social pattern characterization of a user wearing a wearable camera for a long period of time (e.g. a month), 
relying on the visual features transmitted from the captured photo-streams. 
Social pattern characterization is achieved through first, the detection of social interactions of the user and second, their categorization. In the end, different appearances of interacting with the wearer individuals in different social events are localized through face clustering to directly derive the frequency and the diversity of social interactions of the wearer with each individual observed in the images. In the proposed method, social signals for each task are presented in the format of multi-dimensional time-series and LSTM is employed for  the social interaction detection and categorization tasks. A quantitative study over different combination of features for each task is provided, unveiling the impact of each feature on that task.  Evaluation results suggest that in comparison to the frame-level analysis of the social events, sequence-level analysis employing LSTM leads to a higher performance of the model in both tasks. To the best of our knowledge, this is the first attempt at a comprehensive and unified analysis of social patterns of an individual in either ego-vision or third-person vision. This comprehensive study can have important applications in the field of preventive medicine, for example in studying social patterns of patients affected by depression, of elderly people and of trauma survivors. For further details about the proposed methods refer to \cite{aghaei2017all,aghaei2017towards}.

{\small
\bibliographystyle{ieee}
\bibliography{activity}
}

\end{document}